\documentclass[letterpaper]{article} 
\usepackage{aaai23}  
\usepackage{times}  
\usepackage{helvet}  
\usepackage{courier}  
\usepackage[hyphens]{url}  
\usepackage{graphicx} 
\usepackage{natbib}  
\usepackage{caption} 
\usepackage{algorithm}
\usepackage{algorithmic}

\usepackage{newfloat}
\usepackage{listings}
\DeclareCaptionStyle{ruled}{labelfont=normalfont,labelsep=colon,strut=off} 
\floatstyle{ruled}
\newfloat{listing}{tb}{lst}{}
\floatname{listing}{Listing}
\usepackage[caption=false]{subfig}
\usepackage{multirow}
\usepackage{amsmath,bm}
\usepackage{booktabs}
\usepackage{multirow}
\usepackage{amsfonts} 

\usepackage{xcolor}
\usepackage[normalem]{ulem}
\usepackage{soul}

\newcommand{\ie}{\textit{i.e.,}~}
\newcommand{\eg}{\textit{e.g.,}~}
\newcommand{\isa}{\textit{is-a}~}

\title{
DNG: Taxonomy Expansion by Exploring the Intrinsic\\
Directed Structure on Non-gaussian Space
}

\author{
    Songlin Zhai\textsuperscript{\rm 1},
    Weiqing Wang\textsuperscript{\rm 2}\thanks{Corresponding author.},
    Yuanfang Li\textsuperscript{\rm 2},
    Yuan Meng\textsuperscript{\rm 1}
}
\affiliations{

    \textsuperscript{\rm 1} School of Computer Science and Engineering, Southeast University, China\\
    \textsuperscript{\rm 2} Faculty of Information Technology, Monash University, Australia\\
    songlin\_zhai@seu.edu.cn, \{Teresa.Wang,YuanFang.Li\}@monash.edu,  
    yuan\_meng@seu.edu.cn
}

\begin{document}

\maketitle

\begin{abstract}
Taxonomy expansion is the process of incorporating a large number of additional nodes (\ie ``queries'') into an existing taxonomy (\ie ``seed''), with the most important step being the selection of appropriate positions for each query.
Enormous efforts have been made by exploring the seed's structure.
However, existing approaches are deficient in their mining of structural information in two ways: poor modeling of the hierarchical semantics and failure to capture directionality of the \isa relation.
This paper seeks to address these issues by explicitly denoting each node as the combination of inherited feature (\ie structural part) and incremental feature (\ie supplementary part).
Specifically, the inherited feature originates from ``parent'' nodes and is weighted by an inheritance factor.
With this node representation, the hierarchy of semantics in taxonomies (\ie the inheritance and accumulation of features from ``parent'' to ``child'') could be embodied.
Additionally, based on this representation, the directionality of the \isa relation could be easily translated into the irreversible inheritance of features.
Inspired by the Darmois-Skitovich Theorem, we implement this irreversibility by a non-Gaussian constraint on the supplementary feature.
A log-likelihood learning objective is further utilized to optimize the proposed model (dubbed DNG), whereby the required non-Gaussianity is also theoretically ensured.
Extensive experimental results on two real-world datasets verify the superiority of DNG relative to several strong baselines.
\end{abstract}

\section{Introduction}

\begin{figure}[tb]
    \centering
 	\includegraphics[scale=0.95]{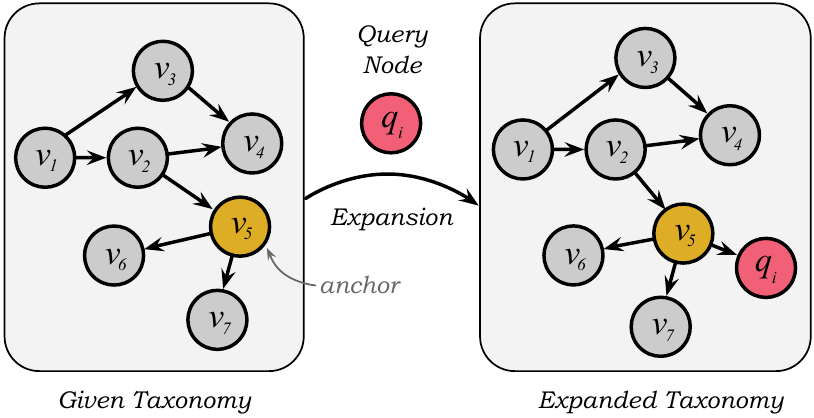}
    \caption{The ``\emph{Given Taxonomy}'' is expanded by appending the ``query'' $q_i$ into the anchor node $v_5$. Specifically, finding appropriate anchors (\eg $v_5$) for a query node (\eg $q_i$) is the crucial step for the taxonomy expansion task.}
    \label{fig:taxonomy-expansion-example}
\end{figure}

Taxonomies provide the hierarchical information for various downstream applications, such as Query Understanding \cite{DBLP:conf/ijcai/WangZWMW15,DBLP:journals/tkde/HuaWWZZ17}, 
Web Search \cite{DBLP:conf/sigmod/WuLWZ12,DBLP:conf/kdd/LiuGNWXLLX19},
and Personalized Recommender System 
\cite{wang2017stsage,huwang2021neuroocomputing}, 
where the \isa relation connects the nodes \cite{DBLP:conf/coling/Hearst92,DBLP:conf/acl/BaiZKCM21}.
However, manually curating a taxonomy is labor-intensive, time-consuming, and difficult to scale, making the automatic taxonomy expansion draw considerable attention \cite{DBLP:conf/www/ShenSX0W020,DBLP:journals/corr/abs-2102048871021,DBLP:journals/corr/abs-2202-04887}.
As depicted in Figure~\ref{fig:taxonomy-expansion-example}, this task aims to insert new nodes (\ie ``query nodes'') into an existing taxonomy (\ie ``seed taxonomy'') by finding appropriate anchors without modifying the seed's original structure for best preserving and utilizing the seed's design.
In this paper, we will also follow this research line to improve taxonomy expansion.

%

Recent works primarily contribute to this task from two directions: capturing the seed's structural information via egonets \cite{DBLP:conf/www/ShenSX0W020}, mini-paths \cite{DBLP:conf/kdd/YuLSFSZ20}, or vertical-horizontal views of a taxonomy \cite{DBLP:journals/corr/abs-2202-04887}; and enhancing node embeddings by implicit edge semantics for a heterogeneous taxonomy \cite{DBLP:conf/www/ManzoorLSL20}, or hyperbolic space \cite{DBLP:conf/acl/AlyAOKBP19,DBLP:conf/emnlp/MaCWP21}.
It has been suggested that exploring the seed's structural information should be more important for this task \cite{DBLP:conf/www/WangZCZL21}.
However, existing works naively incorporate this information by either GNNs \cite{DBLP:journals/ai/YaoLLLC22} or the manually-designed natural language utterance (\eg ``$v_5$ \emph{is a type of} $v_2$'' in Figure~\ref{fig:taxonomy-expansion-example}), leaving two aspects of the seed's structure under-explored, \textbf{Type-\uppercase\expandafter{\romannumeral1}}: the hierarchical semantics among ``parent'' and ``child'' nodes;
and \textbf{Type-\uppercase\expandafter{\romannumeral2}}: the directionality of the \isa relation.
We focus on these two issues in this paper and demonstrate how these two taxonomic structures can be applied to enhance the taxonomy expansion task. 

For further analyzing these two issues, taking the \isa pair (``animal'', ``cat'') as an example, the hypernym ``animal'' is semantically more general than its hyponym ``cat'' \cite{DBLP:conf/eacl/SantusLLW14}, implying two-fold information about these two nodes.
(a) The hyponym ``cat'' contains all features of its hypernym ``animal'', indicating that a part of the feature for ``cat'' could be inherited from ``animal''.
(b) The hypernym ``animal'' does NOT contain all features of its hyponym ``cat'', showing that ``cat'' owns extra features that could not be inherited from ``animal''.
This two-fold information indicates the inclusion relation of semantics between the hypernym and hyponym (\ie ``animal'' is a proper subset of ``cat'' in semantics), reflecting the vertex-level character of a taxonomy.
The relationship also shows the \textbf{inheritance} and \textbf{accumulation} of information from ``parent'' to ``child''.
This variation in information reflects the hierarchical structure of taxonomies (\ie \textbf{Type-\uppercase\expandafter{\romannumeral1}}) and explains why different levels hold different information.

\begin{figure}[tb]
    \centering
 	\includegraphics[scale=1.3]{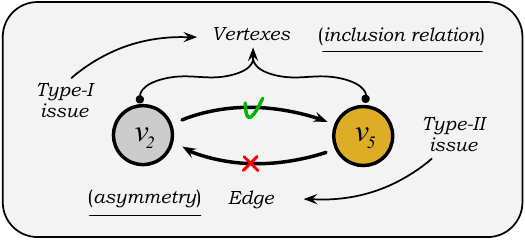}
    \caption{The difference between Type-\uppercase\expandafter{\romannumeral1} and Type-\uppercase\expandafter{\romannumeral2} issues. Type-\uppercase\expandafter{\romannumeral1} focuses on the inclusion relation of node semantics (\ie vertex level), whereas Type-\uppercase\expandafter{\romannumeral2} is designed for the asymmetric transformation of the \isa relation itself (\ie edge level).}
    \label{fig:dng-motivation}
\end{figure}

Besides the hierarchical characteristic, the directionality of the \isa relation (\ie ``cat'' is-a ``animal'', but ``animal'' is-not-a ``cat'') is another meaningful structural information in a taxonomy (\ie \textbf{Type-\uppercase\expandafter{\romannumeral2}}).
This means that the feature inheritance is directed (\eg ``cat'' could inherit features from ``animal'', but not vice versa), which exactly matches the underlying DAG (\underline{D}irected \underline{A}cyclic \underline{G}raph) structure of a taxonomy.
To capture this edge-level character, we seek to derive an asymmetric transformation between hypernym and hyponym features since each relation in a graph could be treated as a mathematical mapping of features.
However, existing approaches \cite{DBLP:conf/eacl/Rimell14,DBLP:conf/acl/FuGQCWL14} focus heavily on manipulating hypernym and hyponym features (\eg resorting to subtraction), which cannot theoretically guarantee the directionality of the \isa relation.

In this paper, we propose the DNG model to address the two types of issues with two coherent techniques, and Figure~\ref{fig:dng-motivation} also illustrates the difference of these two issues.
On one hand, each node is explicitly represented by a combination of the structural feature (inherited from ``parent'') and the supplementary feature (owned by ``itself'' and cannot be inherited from its ``parent''), where the structural feature is weighted by an inheritance factor to measure the inheritability of the node.
Under this representation, the hierarchical semantics is embodied in the feature transmission by inheritance strategy (\ie the feature is passed from one layer to the next) and the feature increment via supplementary mechanism (\ie some extra features will be supplemented in each layer).
On the other hand, benefiting from the improved node representation, the directionality of the \isa relation could be easily converted to the irreversible inheritance of the supplementary feature.
Motivated by the Darmois-Skitovich Theorem, we propose a non-Gaussian constraint on the supplementary feature to implement this irreversibility, therefore theoretically guarantee the relation's directionality.

To optimize DNG, we adopt a log-likelihood learning objective and theoretically prove its effectiveness on modeling structural information.
We conduct extensive experiments on two large real-world datasets from different domains to verify the efficacy of the DNG model.
The experimental results demonstrate that DNG achieves state-of-the-art performance on the taxonomy expansion task.
Additionally, for a better illustration, we will also give a detailed case study about the DNG's ability in modelling taxonomic structure.
To summarize, our major contributions include:
\begin{itemize}
  \item The node representation is improved by combining the structural and supplementary features; thereby, the hierarchical semantics in taxonomies could be explicitly reflected in the modelling process.
  \item A non-Gaussian constraint on the supplementary feature is adopted to theoretically guarantee the directionality of the \isa relation.
  \item A log-likelihood learning objective is employed to optimize the DNG model, ensuring that the required constraint is theoretically guaranteed.
  \item We conduct extensive experiments on two large datasets from different domains to validate the performance of DNG and the efficacy of modeling taxonomic structure.
\end{itemize}

\section{Preliminaries}
\subsection{Task Definition}
\textbf{\emph{Taxonomy}:}
A given taxonomy $\mathcal{T}=\{ \mathcal{V}, \mathcal{E} \}$ can be treated as a directed acyclic graph composed of a vertex set ($\mathcal{V}$) and an edge set ($\mathcal{E}$) \cite{DBLP:conf/www/ShenSX0W020,DBLP:journals/corr/abs-2202-04887}.
An edge $(v_i, v_j) \in \mathcal{E}$ between nodes $v_i$ and $v_j$ represents the \isa relation between them (\eg $v_j$ \isa $v_i$), where $v_i$ and $v_j$ are the hypernym (``\emph{parent}") and the hyponym (``\emph{child}"), respectively.

\noindent \textbf{\emph{Taxonomy Expansion}:}
It refers to the process that appends a set of new nodes (query set $\mathcal{Q}$) into an existing taxonomy (seed taxonomy $\mathcal{T}^0$), whereby a larger taxonomy could be produced.
Hence, this task requires: (1) a seed taxonomy $\mathcal{T}^0=\{ \mathcal{V}^0, \mathcal{E}^0 \}$ with $|\mathcal{V}^0| = N$; and (2) the query set $\mathcal{Q}=\{q_i | 1 \leq i \leq M \}$.
As illustrated in Figure~\ref{fig:taxonomy-expansion-example}, the crucial step is to find appropriate anchors in seed taxonomy $\mathcal{T}^0$ for each query $q_i$.

\subsection{Shannon Entropy}
Shannon entropy is the basic concept for measuring the degree of uncertainty of a random variable.
It is defined for a discrete-valued random variable $X$ as:
\begin{equation}
    H(X) = -\sum p_x(X) \log p_x(X) = -\mathbb{E}\{ \log  p_x(X) \}
\label{eq:entropy-discrete}
\end{equation}

According to Eq.~\ref{eq:entropy-discrete}, entropy for the linear transformation $\bm{y}=\bm{B} \bm{x}$, can be derived as follows \cite{DBLP:books/wi/HyvarinenKO01}:
\begin{equation}
    H(\bm{y})=H(\bm{x})+\ln |\mbox{det }\bm{B}|
\label{eq:entropy-transformation-matrix}
\end{equation}
where det denotes the determinant; $\bm{B}$ and $\ln$ are the mapping matrix and napierian logarithm, respectively.

\section{Methodology}
\subsection{Node Representation}
In order to incorporate the taxonomic structure, each node (\eg $v_i$) is explicitly represented by the combination of structural feature (\emph{inherited from its parents}) and supplementary feature (\emph{could not be inherited from its parents}):
\begin{equation}
    \bm{v}_{i} = \underbrace{\sum_{v_j\in\mbox{PA}_i}  s_{ij} \bm{v}_j}_{\mbox{\emph{structural}}} + \underbrace{\bm{u}_{i}}_{\mbox{\emph{supplementary}}} = \bm{V} \bm{s}_{i} + \bm{u}_{i}
\label{eq:one-node-feature}
\end{equation}
where $\bm{v}_{i}\in\mathbb{R}^d$ and PA$_i$ are the $d$-dimensional representation and the parent set of node $v_i$, respectively.
$s_{ij} \bm{v}_{j}$ denotes the feature inherited from the parent node $v_j$, where the inherited feature is weighted by an inheritance factor $s_{ij}$.
$\bm{u}_{i}$ is the supplementary feature of node $v_i$ which cannot be obtained from its parents. 
$\bm{V}$ indicates the feature matrix of seed's node set ($\mathcal{V}^0$), and $\bm{s}_{i}$ is the inheritance vector of $v_i$ on $\mathcal{V}^0$.

\begin{figure}[tb]
    \centering
 	\includegraphics[scale=0.9]{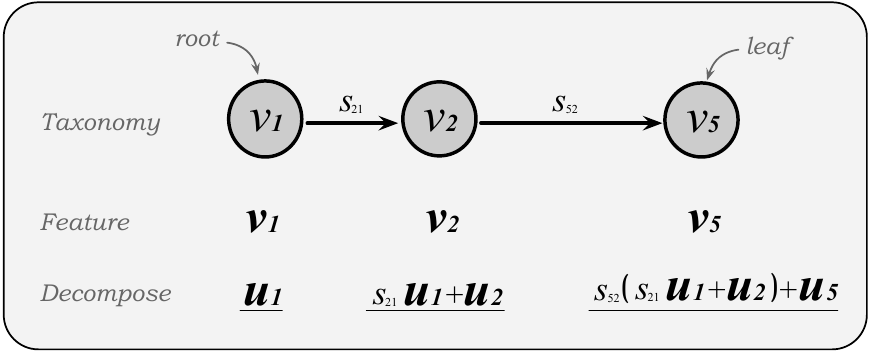}
    \caption{The hierarchical semantics in a taxonomy could be shown by the inheritance and accumulation of features from one layer to the next layer. For node $v_i$, $s_{ij}$ is the inheritance factor to scale the inherited feature from node $v_j$.}
    \label{fig:independence-biase-feature}
\end{figure}

Figure~\ref{fig:independence-biase-feature} illustrates the node representation in a taxonomy fragment with two \isa pairs (\ie $v_5$ \isa $v_2$, and $v_2$ \isa $v_1$).
For the node $v_1$, its feature $\bm{v}_1$ only contains the supplementary part (\ie $\bm{v}_1=\bm{u}_1$) since it is the root and does not inherit any feature from any node.
Node $v_2$ has the parent $v_1$, making its representation $\bm{v}_2$ the combination of $\bm{u}_1$ and $\bm{u}_2$ (\ie $\bm{v}_2 = s_{21} \bm{u}_1 + \bm{u}_2$).
It is observed that $\bm{u}_2$ is independent of $\bm{u}_1$ since neither $\bm{u}_1$ nor $\bm{u}_2$ contains any information about the other.
Similarly, $\bm{v}_5$ can be represented by the inherited feature and its supplementary feature (\ie $\bm{v}_5 = s_{52}(s_{21} \bm{u}_1 + \bm{u}_2) + \bm{u}_5$), from which the independence among $\bm{u}_1$, $\bm{u}_2$, and $\bm{u}_5$ can be also inferred.
Hence, based on Eq.~\ref{eq:one-node-feature}, the hierarchical semantics in a taxonomy could be embodied: the inheritance and accumulation of features from one layer to the next layer.
This example also shows that these supplementary features are independent of each other.

Additionally, the directionality is another crucial attribute of the \isa relation: ``$v_5$ \isa $v_2$, but $v_2$ \textit{is-not-a} $v_5$''.
How to capture this structural information is an issue to be resolved.
In this paper, we attempt to translate this directionality into the irreversible inheritance of features: ``$v_5$ could inherit feature from $v_2$, but $v_2$ could not inherit feature from $v_5$''.
Under this intuition, this attribute could be realized by the irreversible transformation between $\bm{v}_2$ and $\bm{v}_5$ in Eq.~\ref{eq:one-node-feature} (\ie $\bm{v}_2 \neq s_{25} \bm{v}_5 + \bm{u}_2$ under $\bm{v}_5 = s_{52} \bm{v}_2 + \bm{u}_5$).
Inspired by the Darmois-Skitovich Theorem, ensuring the non-Gaussianity of the supplementary feature is a feasible implementation, as proven in the next section.

\subsection{Directionality of \isa on Node Representation}\label{sec:asymmetry-non-gaussian}
Supposing the representation $\bm{v}_i = s_{ij} \bm{v}_j + \bm{u}_i$ admits the inverse version $\bm{v}_j = s_{ji} \bm{v}_i + \bm{u}_j$ under non-Gaussian $\bm{u}_i$, there will exist the following transformation:
\begin{equation}
\begin{aligned}
    \bm{u}_j &= \bm{v}_j - s_{ji} \bm{v}_i\\
    &= \bm{v}_j - s_{ji} (s_{ij} \bm{v}_j + \bm{u}_i) \\
    &= (1 - s_{ji} s_{ij}) \bm{v}_j - s_{ji} \bm{u}_i
\end{aligned}
\label{eq:proof-non-gaussian1}
\end{equation}
As such, the simultaneous equation will be obtained:
\begin{equation}
\left\{
\begin{aligned}
    \bm{v}_i &= s_{ij} \bm{v}_j + \bm{u}_i \\
    \bm{u}_j &= (1 - s_{ji} s_{ij}) \bm{v}_j - s_{ji} \bm{u}_i
\end{aligned}
\right.
\label{eq:proof-non-gaussian2}
\end{equation}
Due to the non-Gaussian $\bm{u}_i$, it can be drawn from Eq.\ref{eq:proof-non-gaussian2} based on the corollary of Darmois-Skitovich Theorem\footnote{See appendix in the arXiv version.} that $\bm{v}_i$ and $\bm{u}_j$ will be dependent (\ie $\bm{v}_i \not \perp \!\!\! \perp \bm{u}_j$).
However, the fundamental condition of the inverse transformation $\bm{v}_j = s_{ji} \bm{v}_i + \bm{u}_j$ is $\bm{v}_i \perp \!\!\! \perp \bm{u}_j$.
This contradiction between the assumption and the deduction verifies that the non-Gaussian supplementary feature will make the reversal node representation given in Eq.~\ref{eq:one-node-feature} false, indicating that the directionality of \emph{is-a} is captured.
In order to ensure this required constraint, we employ a log-likelihood learning target, which will be elaborated in the next section.

\subsection{Learning Target}
\label{sec:learning-target}
Based on the recursive node representation in Eq.~\ref{eq:one-node-feature}, the node set $\mathcal{V}$ can be represented as \footnote{Derivation of Eq.~\ref{eq:node-set-features}: $
\bm{V} = \bm{V} \bm{S} + \bm{U} \Rightarrow
\bm{V} \bm{I} - \bm{V} \bm{S} = \bm{U} \Rightarrow
\bm{V} (\bm{I} - \bm{S}) = \bm{U} \Rightarrow
\bm{V} = \bm{U} (\bm{I} - \bm{S})^{-1}$}:
\begin{equation}
    \bm{V} = \bm{U} \bm{W}^{-1}
\label{eq:node-set-features}
\end{equation}
where $\bm{W} = \bm{I} - \bm{S}$ is dubbed the transition matrix.
$\bm{I}$ and $\bm{S}$ are an identity matrix and the inheritance factor matrix, respectively, and both of them have the size $N \times N$. 
$\bm{U}$ denotes the supplementary feature matrix with the shape of $d \times N$ ($d$ is the dimension of this feature).

From Eq.~\ref{eq:node-set-features}, we can observe that the essential step is to estimate the supplementary feature set $\bm{U}$ and the inheritance factors $\bm{S}$ given the observational feature $\bm{V}$.
For this inheritance factor matrix, the entry with position ($i^{th}, j^{th}$) reflects the probability of node $v_j$ being the ``parent'' of node $v_i$.

In real-word datasets, the feature $\bm{V}$ could be easily obtained from the well-designed pre-trained models \cite{DBLP:journals/corr/abs-1907-11692,DBLP:conf/naacl/DevlinCLT19} by fusing sufficient context information\footnote{DNG focuses on parsing $\bm{U}$ and $\bm{S}$ under the given $\bm{V}$ which can be extracted by other structural approaches with the structural measure, \eg Wu\&P \cite{DBLP:journals/corr/abs-1211-4709}.}.
However, how to learn $\bm{U}$ and $\bm{S}$ from $\bm{V}$ remains to be determined.
To fulfill this goal, we will first analyze the supplementary features $U$ as:
\begin{equation}
    p_{\bm{U}}(\bm{U}) = \prod_{i=1}^{N} p_{\bm{U}}(\bm{u}_i)
\label{eq:error-joint-density}
\end{equation}
where $p_{\bm{U}}(\bm{u}_i)$ is the probability density of $\bm{u}_i$, implemented by $\mbox{tanh}(\bm{u}_i) = (e^{\bm{u}_i} - e^{-\bm{u}_i})/(e^{\bm{u}_i} + e^{-\bm{u}_i})$  \cite{DBLP:books/wi/HyvarinenKO01}.
Then, in order to build the relation between $p_U(\bm{U})$ and $p_V(\bm{V})$, the joint probability density of $\bm{V}$ could be derived from $p_{\bm{U}}(\bm{U})$ by their probability distribution functions (PDFs) \footnote{
First, get the the relation between PDFs: $\mathcal{F}_{\bm{V}}(V) = P_{\bm{V}}(\bm{V} \leq V) = P_{\bm{V}}(\bm{U} \bm{W}^{-1} \leq V) = P_{\bm{U}}(\bm{U} \leq V \bm{W}) = \mathcal{F}_{\bm{U}}(V \bm{W})$.
Then, compute the first-order derivation of PDFs: $p_{\bm{V}}(\bm{V}) = \mathcal{F}_{\bm{V}}^{'}(\bm{V}) = \mathcal{F}_{\bm{U}}^{'}(\bm{V} \bm{W}) = p_{\bm{U}}(\bm{V} \bm{W})~|\mbox{det} \bm{W}|$
}:
\begin{equation}
\begin{aligned}
    p_{\bm{V}}(\bm{V}) 
    = |\mbox{det} \bm{W}| \prod_{i=1}^{N} p_{\bm{U}}(\bm{V} \bm{w}_i)
\end{aligned}
\label{eq:error-feature-density}
\end{equation}
where $\bm{W} = (\bm{w}_i, ..., \bm{w}_n)^{\top}$ indicates how to analyze the supplementary features from the observational representations.

With the definition of $p_{\bm{V}}(\bm{V})$ in hand, the given taxonomy $\mathcal{T}$ could be treated as the joint distribution of the node set $\mathcal{V}$; hence, the learning target can be defined as maximizing the logarithm of the joint probability:
\begin{equation}
\begin{aligned}
    \jmath_{\Theta}(\mathcal{T}) &= 
    -\ln (|\mbox{det} \bm{W}| \prod_{i=1}^{N} p_{\bm{U}}(\bm{V} \bm{w}_i)) \\
    &= -\sum_{i=1}^{N} \ln(p_{\bm{U}}(\bm{V} \bm{w}_i)) - \ln |\mbox{det} \bm{W}|
\end{aligned}
\label{eq:learning-target}
\end{equation}
where $\jmath$ denotes the loss to be minimized, and $\Theta$ is the parameters.
For the DNG model, it is essential to guarantee that the non-Gaussian constraint is satisfied under this optimization, which will be proven in the next section.

\subsection{The Non-Gaussian $\bm{U}$}
To demonstrate that the supplementary feature $\bm{U}$ is a non-Gaussian distribution under the learning objective, let us first derive the expectation of Eq.~\ref{eq:learning-target}:
\begin{equation}
\begin{aligned}
    \mathbb{E}(\jmath) &= -\sum_{i=1}^N \mathbb{E}\{\ln(p_{\bm{U}}(\bm{V} \bm{w}_i))\} - \ln |\mbox{det} \bm{W}| \\
    &= \sum_{i=1}^N H(\bm{V} \bm{w}_i) - \ln |\mbox{det} \bm{W}| \\
    &= \sum_{i=1}^N H(\bm{u}_i) - \ln |\mbox{det} \bm{W}|
\end{aligned}
\label{eq:learning-target-expection}
\end{equation}
Additionally, in information theory, the entropy measures the uncertainty of a random variable; therefore, a variable with more randomness will have higher entropy.
For all types of random variables with the same variance, the Gaussian variable has the highest entropy, leading us to borrow the Negentropy \cite{DBLP:books/wi/HyvarinenKO01} as the measure of a variable's non-Gaussianity.
With the help of Eq.~\ref{eq:learning-target-expection}, the non-Gaussianity of feature $\bm{U}$ can be derived as:
\begin{equation}
\begin{aligned}
    \mbox{NG}(\bm{U}) &= H(\bm{U}_{Gaussian}) - H(\bm{U}) \\
    &= H(\bm{U}_{Gaussian}) + \mathbb{E}\{\ln(p_{\bm{U}}(\bm{V} \bm{W}))\} \\
    &= \mathbb{E}\{\ln(p_{\bm{U}}(\bm{V} \bm{W}))\} - \ln |\mbox{det} \bm{W}| + \\
    & \quad \ H(\bm{U}_{Gaussian}) + \ln |\mbox{det} \bm{W}| \\
    &= -\mathbb{E}(\jmath) + \mbox{\emph{const}}
\end{aligned}
\label{eq:equivalent-negentropy}
\end{equation}
where NG denotes the non-Gaussianity of a variable.
$\bm{U}_{Gaussian}$ is the feature with Gaussian distribution, and $H(\bm{U}_{Gaussian})$ could be treated as a constant \cite{224254ica1993tong,DBLP:books/wi/HyvarinenKO01}.
Additionally, the term $\ln |\mbox{det} \bm{W}|$ could be also considered as a constant since $\mathbb{E}(\bm{V}\bm{V}^{\top}) = 
\bm{W} \mathbb{E}(\bm{U}\bm{U}^{\top}) \bm{W}^{\top}=\bm{I}$ \footnote{$\bm{V}$ is pre-processed to unit variance \cite{DBLP:journals/nn/HyvarinenO00}.} under the unit-variance $\bm{V}$ and $\mbox{det} \bm{I} = 1 = \mbox{det} \bm{W} \mathbb{E}(\bm{U}\bm{U}^{\top}) \bm{W}^{\top} = 
(\mbox{det} \bm{W}) (\mbox{det} \mathbb{E}(\bm{U}\bm{U}^{\top})) (\mbox{det} \bm{W}^{\top})$.
In this transformation, it can draw that the term $\mbox{det} \mathbb{E}(\bm{U}\bm{U}^{\top})$ does not depend on $\bm{W}$, which implies $det \bm{W}$ is a constant  \cite{DBLP:journals/nn/HyvarinenO00}.
Consequently, based on Eq.~\ref{eq:equivalent-negentropy}, it can be also drawn that \emph{maximizing the joint probability of $\bm{V}$ is also equivalent to maximizing the non-Gaussianity of $\bm{U}$.}
This corollary exactly accords with the non-Gaussian condition of the proposed DNG model.

\subsection{DNG Algorithm}
In the DNG model, estimating the matrix $\bm{S}$ is the crucial step for the expansion task.
To launch the learning procedure, the taxonomy with its vertex set, the feature matrix $\bm{V}$ and the edge set are taken as the inputs.
Then, the proposed model DNG is used to analyze the supplementary features and estimate the inheritance factor matrix.
Specially, the gradients of updating the parameters are:
\begin{equation}
\bm{W} \gets \bm{W} + \alpha (\bm{G} \bm{V} + (\bm{W}^{-1})^{\top})
\label{eq:gradient}
\end{equation}
where $\bm{G} = ( 2/sinh(2\bm{u}_1), ..., 2/sinh(2\bm{u}_n) )^{\top}$, and $\alpha$ is the learning rate.
During the inference stage, the inheritance vector $\bm{s}_{q_i}$ for the query $q_i$ is directly computed\footnote{For a more precise estimation, a new iteration is requisite on the whole node set.} based on the query feature $\bm{v}_{q_i}$, the learned supplementary feature matrix and the inheritance matrix.
With $\bm{s}_{q_i}$ in hand, its appropriate positions could be easily found\footnote{Code is available at \url{https://github.com/SonglinZhai/DNG}.}.

\begin{table}
\renewcommand\tabcolsep{10pt}
\centering
\begin{tabular}{lccr}
\toprule[1.0pt]
\textbf{Datasets} &$|\mathcal{V}|$ &$|\mathcal{E}|$ &\#Depth \\
\midrule[0.5pt]
\textbf{MAG-CS} &24,754 &42,329 &6 \\
\textbf{WordNet-Verb} &13,936 &13,408 &13 \\
\bottomrule[1.0pt]
\end{tabular}
\caption{Statistics of datasets, where $|\mathcal{V}|$  and $|\mathcal{E}|$ are the number of nodes and edges in taxonomies, respectively. \#Depth indicates the depth of taxonomies.}
\label{tab:statistics-taxonomy}
\end{table}

\section{Experiments}
\subsection{Experimental Setup}
\subsubsection{Datasets}
We evaluate DNG and compare models on two large public real-world taxonomies:
\begin{itemize}
    \item \textbf{Microsoft Academic Graph (MAG)} is a public Field-of-Study (FoS) taxonomy \cite{DBLP:conf/www/SinhaSSMEHW15}.
    This taxonomy contains over 660,000 scientific nodes and more than 700,000 \isa relations.
    Following \cite{DBLP:conf/www/ShenSX0W020,DBLP:conf/aaai/ZhangSZCSM021,DBLP:journals/corr/abs-2202-04887}, we re-construct a sub-taxonomy from MAG by picking the nodes of the computer science domain, named \textbf{MAG-CS}.
    Additionally, the node embeddings  extracted via word2vec \cite{DBLP:conf/nips/MikolovSCCD13} based on the related paper abstracts corpus are used as the initial embeddings.
    \item \textbf{WordNet} is another large taxonomy, containing various concepts \cite{DBLP:conf/semeval/JurgensP16}. Following \cite{DBLP:conf/aaai/ZhangSZCSM021,DBLP:journals/corr/abs-2202-04887}, the concepts of the verb sub-taxonomy are selected based on WordNet 3.0, referred as \textbf{WordNet-Verb} since this sub-field is the part that has been fully-developed in WordNet.
    Analogously, the fasttext embeddings\footnote{The wiki-news-300d-1M-subword.vec.zip on official website is used.} are also generated as the initial feature vectors \cite{DBLP:conf/www/ShenSX0W020,DBLP:conf/aaai/ZhangSZCSM021}.
    In addition, a pseudo root named ``verb" is inserted as the root node to generate a complete taxonomy because there are isolated components in the constructed taxonomy.
\end{itemize}
Following the  dataset splits used in \cite{DBLP:conf/aaai/ZhangSZCSM021,DBLP:journals/corr/abs-2202-04887}, 1,000 nodes are randomly chosen to test our model in each dataset.
The statistical information for these two datasets is listed in Table~\ref{tab:statistics-taxonomy}.

\begin{table}
\renewcommand\arraystretch{1.2}
\centering
\begin{tabular}{|l|c|c|c|r|}
\hline
\multirow{2}{*}{\textbf{Models}} &\multicolumn{4}{|c|}{\textbf{MAG-CS}} \\
\cline{2-5}
&{Recall} &{Precision} &MR &MRR \\
\cline{1-5}
{TaxoExpan} &0.100 &0.163 &197.776 &0.562 \\
{ARBORIST} &0.008 &0.037 &1142.335 &0.133 \\
{TMN} &0.174 &0.283 &118.963 &0.689 \\
{GenTaxo} &0.149 &0.294 &140.262 &0.634 \\
{TaxoEnrich} &0.182 &0.304 &67.947 &0.721 \\
\cline{1-5}
\textbf{DNG} &\textbf{0.268} &\textbf{0.317} &\textbf{59.354} &\textbf{0.754} \\
\hline
\hline
\multirow{2}{*}{\textbf{Models}} &\multicolumn{4}{|c|}{\textbf{WordNet-Verb}} \\
\cline{2-5}
&{Recall} &{Precision} &MR &MRR \\
\cline{1-5}
{TaxoExpan} &0.085 &0.095 &665.409 &0.406 \\
{ARBORIST} &0.016 &0.067 &608.668 &0.380 \\
{TMN} &0.110 &0.124 &615.021 &0.423 \\
{GenTaxo} &0.094 &0.141 &6046.363 &0.155 \\
{TaxoEnrich} &0.162 &\textbf{0.294} &217.842 &0.481 \\
\cline{1-5}
\textbf{DNG} &\textbf{0.199} &0.201 &\textbf{43.723} &\textbf{0.513} \\
\hline
\end{tabular}
\caption{Performance of all compared models, where ``Recall'' and ``Precision'' refer to the top-1 evaluation metrics. The results of baselines are from \citet{DBLP:journals/corr/abs-2202-04887}.}
\label{tab:performances}
\end{table}

\subsubsection{Compared Models}
Five recent methods are adopted to verify the performance of DNG model:
\begin{itemize}
    \item[(1)] \textbf{TaxoExpan} \cite{DBLP:conf/www/ShenSX0W020} employs a position-enhanced graph neural network to capture the taxonomic structure based on the InfoNCE \cite{DBLP:journals/corr/abs-1807-03748} for better model robustness.
    \item[(2)] \textbf{ARBORIST} \cite{DBLP:conf/www/ManzoorLSL20} aims to resolve the issues in heterogeneous taxonomies via a large margin ranking loss and a dynamic margin function.
    \item[(3)] \textbf{TMN} \cite{DBLP:conf/aaai/ZhangSZCSM021} performs the taxonomy completion task by computing the matching score of the triple (query, hypernym, hyponym). We employ the model for the expansion task in this paper.
    \item[(4)] \textbf{GenTaxo} \cite{DBLP:conf/kdd/ZengLYC021} resorts to the sentence-based and subgraph-based node encodings to perform the matching of each query node. Following \cite{DBLP:journals/corr/abs-2202-04887}, the GenTaxo++ is adopted in this paper since the original framework contains a name generation module which is not the focus of this paper.
    \item[(5)] \textbf{TaxoEnrich} \cite{DBLP:journals/corr/abs-2202-04887} achieves state-of-the-art performance on the taxonomy expansion task. This approach leverages semantic and structural information to better represent nodes, which is a strong baseline in this paper.
\end{itemize}

\subsubsection{Evaluation Metrics}
A ranked list of nodes is provided to choose \textcolor{black}{appropriate} anchors for each query.
Following \cite{DBLP:conf/www/ShenSX0W020,DBLP:conf/aaai/ZhangSZCSM021,DBLP:journals/corr/abs-2202-04887}, four rank-based metrics are used to evaluate the performance of DNG and the compared methods:
\begin{itemize}
    \item \textbf{Precision}\emph{@k} is defined as $$\mbox{Precison@}k=\#\mbox{TP@}k/k$$
    where $\#\mbox{TP@}k$ is 1 if the ground-truth node is in the top-$k$ predictions, otherwise it is 0. The results reported in Table~\ref{tab:performances} are the mean value of all test cases on Precision@1.
    \item \textbf{Recall}\emph{@k} is computed by: $$\mbox{Recall@}k=\#\mbox{TP@}k/\#\mbox{GT}$$
    where $\#\mbox{GT}$ is the number of ground truth labels for one case. The results reported in Table~\ref{tab:performances} are also the mean value of all test cases on Recall@1.
    \item \textbf{Mean Rank} (MR) gives the average rank of true position among all candidate positions.
    Following \cite{DBLP:conf/aaai/ZhangSZCSM021,DBLP:journals/corr/abs-2202-04887}, for multi-anchor queries, we also first compute the result of each individual triplet and then take the average.
    \item \textbf{Mean Reciprocal Rank} (MRR) defines the reciprocal rank of the true position for each query.
    In this paper, the scaled MRR is also adopted to amplify the difference among different approaches \cite{DBLP:conf/www/ShenSX0W020,DBLP:conf/aaai/ZhangSZCSM021,DBLP:journals/corr/abs-2202-04887}.
\end{itemize}

\subsection{Performance Evaluation}
Similar to \cite{DBLP:conf/www/ShenSX0W020,DBLP:conf/aaai/ZhangSZCSM021,DBLP:journals/corr/abs-2202-04887}, we also split the original taxonomy into separate sub-taxonomies, with each sub-taxonomy containing all paths from a source node (\eg the root node) to a destination node (\eg one leaf node).
During inference, candidate anchors are first initially produced from these distinct sub-taxonomies.
Then these candidate anchors are recombined by combining the anchors from the nearest sub-taxonomies to generate better anchors.
After several iterations, the ultimate predictions could be generated.

Table~\ref{tab:performances} summarizes the performance of all compared models on Precision, Recall, MR and MRR evaluation metrics.
Generally, the DNG model outperforms all baseline approaches on most metrics and especially gains a large margin improvement in Recall across all datasets, indicating the actual anchors being ranked at high positions.
Additionally, it has been shown that not all models are capable of producing comparatively superior results for all evolution metrics.
Despite the findings demonstrating the excellent performance of TaxoEnrich, the proposed DNG model still surpasses this strong baseline on the majority of criteria.
The advantage of the DNG model on MR and MRR reveals its superior ranking performance, validating the effectiveness of the designed node representation and the non-Gaussian constraint presented in this paper.

\begin{figure}[tb]
    \centering
 	\includegraphics[width=0.9\columnwidth]{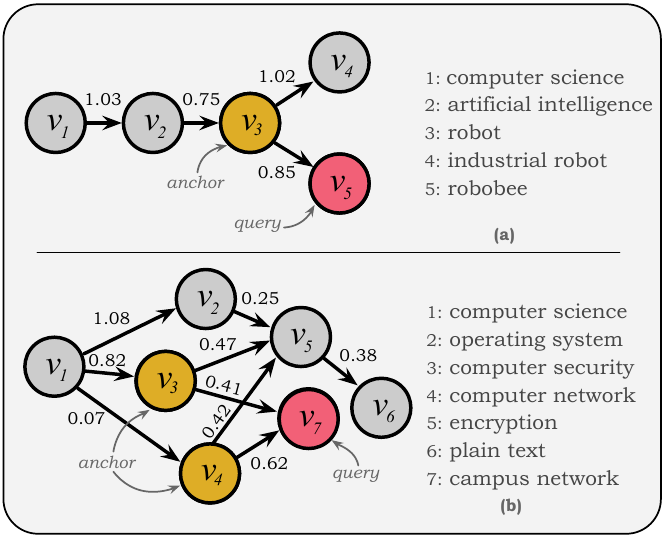}
    \caption{Two examples for validating DNG's ability to capture the hierarchical semantics. The upper figure depicts the expansion of a query into a taxonomy fragment with a chain structure. However, the bottom one is a slightly more complicated example with a graph-structure taxonomy.}
    \label{fig:case-study-magcs}
\end{figure}

\subsection{Capturing Hierarchical Semantics}
Section~\ref{fig:case-study-magcs} presents a re-designed node representation to capture the hierarchical semantics reflected by the inheritance factor.
By estimating $\bm{S}$, \textcolor{black}{appropriate} anchors could be determined by locating the \textcolor{black}{nodes with high factors}, which will be illustrated in this section.

Figure~\ref{fig:case-study-magcs} gives two real cases of the MAG-CS dataset for validating DNG's ability to capture taxonomic structure.
In the top sub-figure, there are five nodes, each with a single parent in the taxonomy, where $v_1 \to v_2 \to v_3 \to v_4$ and $v_5$ are the candidate nodes and the query node, respectively.
Hence, the DNG model needs to estimate the inheritance vector $\bm{s}_5$ on these candidate nodes.
After carrying out the inference, the estimated $\bm{s}_5$ is $[0, 0, 0.8519, 0, 0]$, indicating $v_3$ is its only ``parent'' with the inheritance factor $0.8519$ and it does not inherit any feature from any other nodes.
Since the inheritance factor of $v_3$ is the largest and others are $0$, the appropriate anchor of query $v_5$ is $v_3$.
Figure~\ref{fig:case-study-magcs} (b) depicts a more complicated example in which the convoluted relationships could more convincingly demonstrate DNG's ability.
The matching process for the query node $v_7$ is to estimate $\bm{s}_7$ on other nodes, and the result is $[0, 0, 0.4116, 0.6156, 0, 0, 0]$, indicating $v_7$ inherits features from both of $v_3$ and $v_4$.
Similarly, appropriate anchors are $v_3$ and $v_4$ since all other factors are equal to zero.
Another notable finding is that $v_7$ (``campus network'') inherits more features from $v_4$ (``computer network'') than from $v_3$ (``computer security''), despite the fact that both $v_3$ and $v_4$ are the ``parents''. This is consistent with their ground-truth relations in the MAG-CS dataset, which further validates the DNG's superiority in identifying appropriate anchors for queries in taxonomic structures.

\begin{figure}[tb]
    \centering
 	\includegraphics[width=0.75\columnwidth]{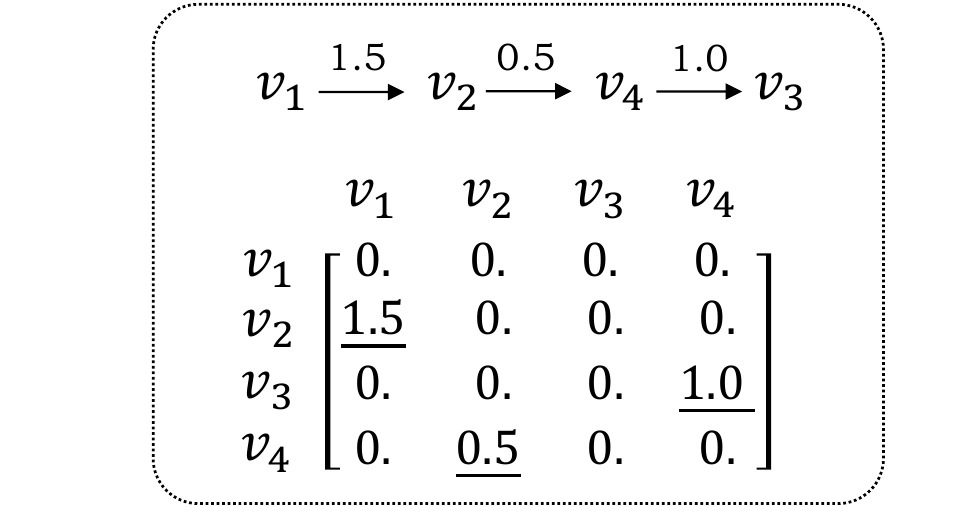}
    \caption{A toy taxonomy with four nodes and three edges. The bottom sub-figure depicts the actual inheritance matrix, where the entity in row $i^{th}$ and column $j^{th}$ denotes the inheritance factor used to weight the inherited feature for the node $v_i$. These underlined entries indicate that actual connections exist at the current locations. For instance, the underlined factor $1.5$ with location ($2^{nd}, 1^{st}$) shows that the connection $v_1 \to v_2$ is a ground-truth one; however, the entity $0.$ with location ($1^{st}, 2^{nd}$) is the opposite.}
    \label{fig:direction-inheritance-mx}
\end{figure}

\subsection{Capturing Directionality and Taxonomic Structure}
One fundamental contribution of this paper is capturing the directionality of the \isa relation under the non-Gaussian constraint; thereby, the structure of a taxonomy could be illustrated. 
Theoretically, to validate this, we need to solve the exact backward equation (\ie $\bm{v}_j=f(\bm{v}_i)$) under the non-Gaussian constraint to see if it obeys the reversed form of Eq.~\ref{eq:one-node-feature} (\ie $\bm{v}_{j} = \sum_{v_i\in\mbox{PA}_j}  s_{ji} \bm{v}_i + \bm{u}_{j}$).
However, despite the fact that the forward equation (see Eq.~\ref{eq:one-node-feature}) has been theoretically proven correct, it is still difficult to determine the exact backward equation due to the lack of its prior knowledge.
Consequently, we devise an alternative synthetic experiment to serve a similar purpose.
Specifically, on the one hand, the distance between the actual and predicted features is also a measure of model's reliability, with a lower value indicating better performance.
On the other hand, supposing the backward equation obeys the reversed Eq.~\ref{eq:one-node-feature}, the difference between the distances computed by the forward and backward equations could reflect the ``deviation degree'' of the backward one.
It is anticipated that this ``deviation degree'' will have a larger value for nodes with non-Gaussian supplementary features compared with that for Gaussian ones. We use a uniform distribution as a non-Gaussian distribution.
Consequently, these distance differences of a uniformly-distributed and a Gaussianly-distributed supplementary features are computed to evaluate the reasonableness of the hypothetical backward equation in this section.

\begin{figure}[tb]
    \centering
 	\includegraphics[width=0.75\columnwidth]{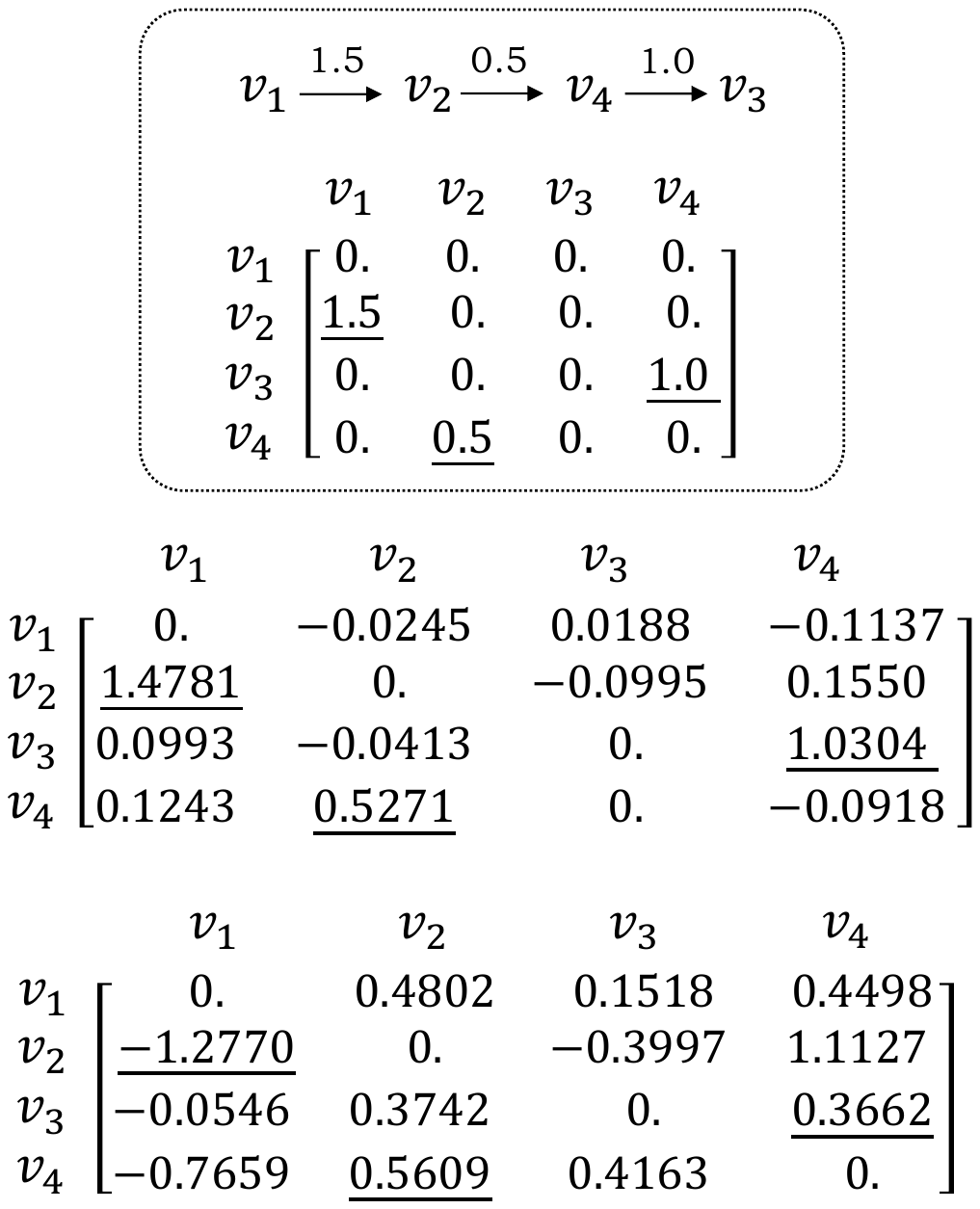}
    \caption{The estimated inheritance matrices of the uniformly-distributed (the top sub-figure) and Gaussianly-distributed (the bottom one) supplementary feature matrices with respect to the nodes in Figure~\ref{fig:direction-inheritance-mx}. It is observed that the value of inheritance matrix under the Gaussian distribution diverge significantly from the actual value (in Figure~\ref{fig:direction-inheritance-mx}),  and the structure under this distribution is also erroneous.}
    \label{fig:direction-inheritance-mx-predict}
\end{figure}

To conduct the experiment described above, two virtual taxonomies with four nodes and an identical structure (see Figure~\ref{fig:direction-inheritance-mx}) are constructed using a uniform and a Gaussian distribution, respectively, where the edge set is $\{v_1 \to v_2 \to v_4 \to v_3\}$ with inheritance factors $\{1.5, 0.5, 1.0\}$.
The DNG model is then employed to estimate the matrix $\bm{S}$ and the supplementary feature $\bm{U}$ under these two distributions.
For the uniform distribution, the distance between the actual and the predicted features with forward equation is: $\{0.51, 0.58, 0.50, 0.59\}$, while the distance between the actual and the backward-predicted features is $\{0.58, 1.25, 1.63, 1.13\}$.
An analogous procedure is followed to obtain the distances for the Gaussian distribution: $\{0.79, 0.84, 0.83, 0.83\}$ (forward) and $\{0.79, 1.47, 1.44, 1.11\}$ (backward).
%
%
%
As a result, the mean ratios of the ``deviation degree'' for the uniform and Gaussian distributions are 107.19\% and 45.56\%, respectively.
The result implies that the hypothetical backward equation incurs a higher loss for the uniform distribution, demonstrating the irreversibility of Eq.~\ref{eq:one-node-feature} under the non-Gaussian constraint.

Apart from demonstrating DNG's ability in capturing the directionality, we also conduct experiments to validate its ability in capturing the entire taxonomic structure. 
Figure~\ref{fig:direction-inheritance-mx-predict} presents the estimated inheritance matrices $\bm{S}$ of the taxonomy (see Figure~\ref{fig:direction-inheritance-mx}) under different distributions.
As illustrated in the upper matrix, compared to the estimated $\bm{S}$ with Gaussian distribution, the one with non-Gaussian constraint (the top figure) more precisely capture the taxonomic structure, since the ground-truth ``parents'' have the larger inheritance factor, validating DNG's ability in modeling taxonomic structure.

\section{Related Work}

To improve the taxonomy expansion task,
enormous efforts have been made to incorporate the structural information of the seed.
\citet{DBLP:conf/www/ManzoorLSL20} learns node embeddings with multiple linear maps to more effectively integrate the structural information of implicit edge semantics for heterogeneous taxonomies.
For the specific modelling structural information, \citet{DBLP:conf/www/ShenSX0W020} employs the egonet with a position-enhanced graph neural network, and a training scheme based on the InfoNCE loss is also employed to enhance the noise-robustness.
However, \citet{DBLP:conf/kdd/YuLSFSZ20} converts the expansion task into the matching between query nodes and mini-paths with anchors to capture the locally structural information.
For optimizing the taxonomic structure, \citet{DBLP:journals/corr/abs-2102048871021} is the only one that considers the relationship among the different queries to ensure that the generated taxonomy has the optimal structure.
\citet{DBLP:conf/aaai/ZhangSZCSM021} adds a new node by simultaneously considering the hypernym-hyponym pair, thereby overcoming the limitations of hyponym-only mode.
Additionally, \citet{DBLP:conf/www/WangZCZL21} aims to maximize the coherence of the expanded taxonomy in order to take advantage of the hierarchical structure of the taxonomy.
\citet{DBLP:journals/corr/abs-2202-04887} combines the semantic features (extracted by surface names) and structural information (generated by the pseudo sentences of both vertical and horizontal views) to better represent each node, achieving state-of-the-art performance on this task.

\begin{figure}[tb]
    \centering
 	\includegraphics[width=0.485\columnwidth]{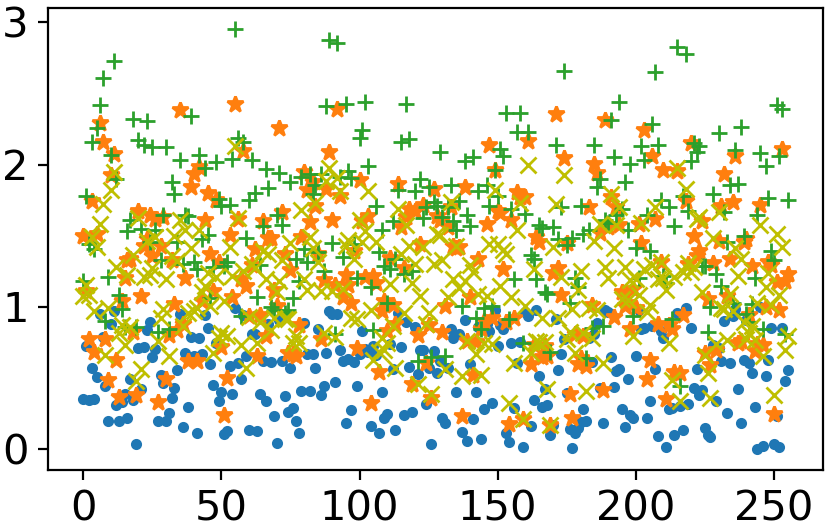} \ 
 	\includegraphics[width=0.42\columnwidth]{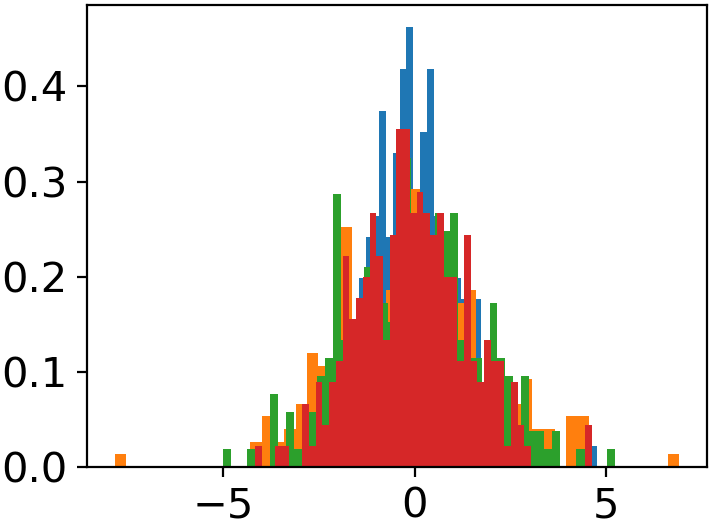}
    \caption{Visualizations of the nodes (in Figure~\ref{fig:direction-inheritance-mx}) under the uniform (left one) and Gaussian (right one) distributions.}
    \label{fig:direction-dists}
\end{figure}
%


\section{Conclusion}
In this paper, we sought to improve the taxonomy expansion task by investigating the taxonomic structure from hierarchical semantics and the directionality of the \isa relation, by which the DNG model was proposed.
On the one hand, to model hierarchical semantics, each node is represented by the combination of the structure and the supplementary features, with the structure part weighted by an inheritance factor.
On the other hand, based on the node representation, the \isa's directionality is easily translated into the irreversible inheritance of features, implemented by the non-Gaussian distribution of the supplementary features.
In order to optimize DNG, we employed the log-likelihood learning objective and demonstrated the equivalency between this learning target and the required constraint.
Finally, extensive experiments were conducted on two large real-world datasets, and the results not only verified the DNG's superiority over other compared models on the expansion task but also confirmed its ability to capture the taxonomic structure.

\appendix
\section{DNG Optimization}
\label{sec:optimization}
To optimize the DNG model, the derivation of Eq.~\ref{eq:learning-target} with regard to $w_{ij}$ is required based on SGD \cite{Bottou91stochasticgradient}:
\begin{equation}
\begin{aligned}
\frac{\partial -\jmath}{\partial w_{ij}} &= \sum\nolimits_{k=1}^{N} 1/p_U (\bm{V} \bm{w}_k) \frac{\partial p_U (\bm{V} \bm{w}_k)}{\partial w_{ij}} + \frac{1}{\mbox{det} \bm{W}} \frac{\partial ~\mbox{det} \bm{W}}{\partial w_{ij}} \\
&= 1/p_U (\bm{V} \bm{w}_i) \frac{\partial p_U (\bm{V} \bm{w}_i)}{\partial w_{ij}} + \frac{1}{det \bm{W}} (-1)^{i+j} M_{ij} \\
\end{aligned}
\end{equation}
where $(-1)^{i+j} M_{ij}$ is the algebraic complement of $w_{ij}$.
The distribution of supplementary feature is implemented by $\mbox{tanh}(\cdot)$ in this paper,
making 
$1/p_U (\bm{V} \bm{w}_i) = \frac{e^{\bm{V} \bm{w}_i} - e^{-{\bm{V} \bm{w}_i}}}{e^{\bm{V} \bm{w}_i} + e^{-{\bm{V} \bm{w}_i}}}$.
Additionally, $\frac{\partial p_U (\bm{V} \bm{w}_i)}{\partial w_{ij}}$ could be also computed as: 
\begin{equation}
\begin{aligned}
&\frac{\partial \{ (e^{\bm{V} \bm{w}_i} - e^{-{\bm{V} \bm{w}_i}})/(e^{\bm{V} \bm{w}_i} + e^{-{\bm{V} \bm{w}_i}}) \}}{\partial w_{ij}} \frac{\partial (\bm{V} \bm{w}_i)}{\partial w_{ij}} \\
&\quad = \frac{(e^{\bm{V} \bm{w}_i} + e^{-{\bm{V} \bm{w}_i}})^2 - (e^{\bm{V} \bm{w}_i} - e^{-{\bm{V} \bm{w}_i}})^2}{(e^{\bm{V} \bm{w}_i} + e^{-{\bm{V} \bm{w}_i}})^2} \bm{v}_j \\
&\quad = \bm{v}_j/(\frac{e^{\bm{V} \bm{w}_i} + e^{-{\bm{V} \bm{w}_i}}}{2})^2 = \bm{v}_j/cosh^2(\bm{V} \bm{w}_i)
\end{aligned}
\end{equation}
where $\bm{v}_j$ is the $j^{th}$ column of $\bm{V}$.
Then, the derivation can be converted to:
\begin{equation}
\begin{aligned}
\frac{\partial -\jmath}{\partial w_{ij}}
&= \bm{v}_j/(tanh(\bm{V} \bm{w}_i) cosh^2(\bm{V} \bm{w}_i)) + \frac{1}{\mbox{det} \bm{W}} (-1)^{i+j} M_{ij} \\
&= 2\bm{v}_j/sinh(2\bm{V} \bm{w}_i) + \frac{1}{\mbox{det} \bm{W}} (-1)^{i+j} M_{ij}
\end{aligned}
\end{equation}
Let $\bm{G} = ( 2/sinh(2\bm{V} \bm{w}_1), ..., 2/2sinh(\bm{V} \bm{w}_n) )^{\top}$, the derivation could be converted to:
\begin{equation}
\frac{\partial - \jmath}{\partial w_{ij}} = \bm{g}_i \bm{v}_j + \frac{1}{\mbox{det} \bm{W}} (-1)^{i+j} M_{ij}
\end{equation}

Since $\bm{W}^*$ is the adjoint matrix of $\bm{W}$, the entry of $\bm{W}^*$ in the $i^{th}$ row and $j^{th}$ column is the algebraic complement of $w_{ij}$ ($(-1)^{i+j} M_{ij}$).
Due to $\bm{W} \bm{W}^* = (\mbox{det} \bm{W}) \bm{I}$, the derivation could be translated into:
\begin{equation}
\begin{aligned}
\frac{\partial - \jmath}{\partial \bm{W}} = \bm{G} \bm{V} + \frac{(\bm{W}^*)^{\top}}{\mbox{det} \bm{W}} &= \bm{G} \bm{V} + \frac{((\mbox{det} \bm{W}) \bm{W}^{-1})^{\top}}{\mbox{det} \bm{W}} \\
&= \bm{G} \bm{V} + (\bm{W}^{-1})^{\top}
\end{aligned}
\end{equation}
Finally, the gradients of updating parameters are:
\begin{equation}
\bm{W} \gets \bm{W} + \alpha (\bm{G} \bm{V} + (\bm{W}^{-1})^{\top})
\end{equation}


\section{The Darmois-Skitovich Theorem}\label{sec:Darmois-Skitovich-theorem}

According to the Darmois-Skitovich Theorem \cite{Darmois-Skitovich1,Darmois-Skitovich2},
let $\{ X_1, X_2, ..., X_n \}$ be independent, non-degenerate random variables: if there exist coefficients $\alpha_1, ..., \alpha_n$ and $\beta_1, ..., \beta_n$ that are all non-zero, the following two linear combinations
\begin{equation}
\begin{aligned}
    A &= \alpha_1 X_1 + \alpha_2 X_2 + ... + \alpha_n X_n \\
    B &= \beta_1 X_1 + \beta_2 X_2 + ... + \beta_n X_n
\end{aligned}
\end{equation}
are independent, then each random variable $X_i$ will be normally distributed.
The contrapositive of the special case of this theorem for two random variables ($n = 2$) will be used in the paper (\ie the following corollary).

\noindent \textbf{Corollary}: If either of the independent random variables $X_1$ and $X_2$ are non-Gaussian, the following two linear combinations
\begin{equation}
\begin{aligned}
    A = \alpha_1 X_1 + \alpha_2 X_2, \quad
    B = \beta_1 X_1 + \beta_2 X_2
\end{aligned}
\end{equation}
will not be independent (\ie $A$ and $B$ must be dependent).

\section{The Independence in $\bm{U}$}
The joint probability density shown in Eq.~\ref{eq:error-joint-density} is based on the independence of supplementary features.
Analogously, to prove the equivalency between DNG optimization and the required independence, we should firstly review the mutual information:
\begin{equation}
I(x_1, x_2, ..., x_n) = \sum\nolimits_{i=1}^{n} H(x_i) - H(\bm{x})
\label{eq:mutual-information}
\end{equation}
where $\bm{x}$ is a vector contains a set of variables $x_i$.

According to Eq.~\ref{eq:mutual-information}, the mutual information of $\bm{U}$ could be computed as:
\begin{equation}
    I(\bm{u}_1, \bm{u}_2, ..., \bm{u}_n) = \sum\nolimits_{i=1}^{n} H(\bm{u}_i) - H(\bm{U})
\label{eq:mutual-information-source}
\end{equation}

By applying Eq.~\ref{eq:entropy-transformation-matrix}, the mutual information shown in Eq.\ref{eq:mutual-information-source} will be transformed into:
\begin{equation}
    I(\bm{u}_1, \bm{u}_2, ..., \bm{u}_n) = \sum\nolimits_{i=1}^{n} H(\bm{u}_i) - H(\bm{V}) - \ln |\mbox{det} \bm{W}|
\label{eq:mutual-information-source2}
\end{equation}

Based on Eq.~\ref{eq:mutual-information-source2} and Eq.~\ref{eq:learning-target-expection}, the relation between the mutual information and the learning target could be denoted as:
\begin{equation}
    \mathbb{E}(\jmath) = I(\bm{u}_1, \bm{u}_2, ..., \bm{u}_n) + \mbox{\emph{const}}
\label{eq:equivalent-mutual-information}
\end{equation}
where \emph{const} denotes a constant.
From Eq.~\ref{eq:equivalent-mutual-information}, it can be drawn that \emph{maximizing the joint probability of $\bm{V}$ is equivalent to minimizing the mutual information of $\bm{U}$.}
This conclusion accords with the prerequisite of Eq.~\ref{eq:error-joint-density}.

\section*{Acknowledgements}
This material is based on research sponsored by Defense Advanced Research Projects Agency (DARPA) under agreement number HR0011-22-2-0047. 
The U.S. Government is authorised to reproduce and distribute reprints for Governmental purposes notwithstanding any copyright notation thereon. 
The views and conclusions contained herein are those of the authors and should not be interpreted as necessarily representing the official policies or endorsements, either expressed or implied, of DARPA or the U.S. Government.

\bibliography{Zhai}

\end{document}